\documentclass[a4paper]{styles/svproc}
\usepackage[suppress]{color-edits}
\usepackage{hyperref}
\addauthor{ms}{magenta}
\addauthor{pw}{teal}
\addauthor{tc}{blue}
\addauthor{ts}{pink}

\usepackage{graphicx,psfrag,amsmath,amsfonts,verbatim,tikz}
\usepackage{booktabs}
\usepackage{xcolor}
\usepackage{hyperref}
\usepackage{tikz}
\usepackage{listings}
\usepackage{bbm}
\usepackage{fancyvrb}
\usepackage[most]{tcolorbox}
\usepackage{multirow}
\usepackage[ruled,vlined]{algorithm2e}
\usepackage{algorithmicx}
\usepackage{amssymb}
\usepackage{pifont}
\newcommand{\cmark}{\ding{51}}%
\newcommand{\xmark}{\ding{55}}%

\newcommand{\inR}{\in \mathbb{R}}
\newcommand{\CFree}{\mathcal{C}^\mathrm{free}}
\newcommand{\Cfree}{\CFree}

\newcommand{\minz}{\mathop{\textbf{minimize} }}

\newcommand{\calA}{\ensuremath{\mathcal{A}}}
\newcommand{\calB}{\ensuremath{\mathcal{B}}}
\newcommand{\calC}{\ensuremath{\mathcal{C}}}
\newcommand{\calD}{\ensuremath{\mathcal{D}}}
\newcommand{\calE}{\ensuremath{\mathcal{E}}}

\newcommand{\calG}{\ensuremath{\mathcal{G}}}
\newcommand{\calH}{\ensuremath{\mathcal{H}}}

\newcommand{\calO}{\ensuremath{\mathcal{O}}}
\newcommand{\calP}{\ensuremath{\mathcal{P}}}

\newcommand{\calR}{\ensuremath{\mathcal{R}}}
\newcommand{\calS}{\ensuremath{\mathcal{S}}}

\newcommand{\subjectto}{\mathop{\textbf{subject to}}}

\newcommand{\Prob}{\mathop{\bf Pr}}

\usepackage[utf8]{inputenc}

\DeclareFixedFont{\ttb}{T1}{txtt}{bx}{n}{10} 
\DeclareFixedFont{\ttm}{T1}{txtt}{m}{n}{10}  

\usepackage{color}
\definecolor{deepblue}{rgb}{0,0,0.5}
\definecolor{deepred}{rgb}{0.6,0,0}
\definecolor{deepgreen}{rgb}{0,0.5,0}

\usepackage{listings}

\definecolor{codegreen}{rgb}{0,0.6,0}
\definecolor{codegray}{rgb}{0.5,0.5,0.5}
\definecolor{codepurple}{rgb}{0.58,0,0.82}
\definecolor{backcolour}{rgb}{0.95,0.95,0.92}

\lstdefinestyle{mystyle}{
	backgroundcolor=\color{backcolour},   commentstyle=\color{codegreen},
	keywordstyle=\color{magenta},
	numberstyle=\tiny\color{codegray},
	stringstyle=\color{codepurple},
	basicstyle=\ttfamily\footnotesize,
	breakatwhitespace=false,         
	breaklines=true,                 
	captionpos=b,                    
	keepspaces=true,                 
	numbers=left,                    
	numbersep=5pt,                  
	showspaces=false,                
	showstringspaces=false,
	showtabs=false,                  
	tabsize=2
}
\usepackage[abbreviations]{glossaries-extra}
\newabbreviation{gcs}{GCSTrajOpt}{\emph{Graph of Convex Sets}}
\usepackage{cleveref}
\usepackage{pgfplots}
\RequirePackage{pgfplots}
\pgfplotsset{compat=1.17}
\usepackage{subcaption}
\usepackage{placeins}

\crefname{section}{sec.}{secs.}
\Crefname{section}{Sec.}{Secs.}
\crefname{figure}{fig.}{figs.}
\Crefname{figure}{Fig.}{Figs.}
\crefname{algorithm}{alg.}{algs.}
\Crefname{algorithm}{Alg.}{Algs.}
\crefname{table}{tab.}{tabs.}
\Crefname{table}{Tab.}{Tabs.}

\title{Faster Algorithms for Growing Collision-Free Convex Polytopes in Robot Configuration Space}
\titlerunning{Faster Algorithms for Growing Large Collision-Free Convex Polytopes} 
\authorrunning{Werner et al.}
\institute{Massachusetts Institute of Technology, Cambridge, MA 02139\\ \texttt{\{wernerpe,tcohn,rhjiang,tseyde,msimchow,russt,rus\}@mit.edu}}

\author{Peter Werner \and Thomas Cohn$^\star$ \and Rebecca H. Jiang\thanks{Denotes equal contribution.} \and Tim Seyde \and Max Simchowitz \and Russ Tedrake \and Daniela Rus}

\begin{document}
\maketitle
\vspace{-0.5cm}
\begin{abstract}
We propose two novel algorithms for constructing convex collision-free polytopes in robot configuration space. 
Finding these polytopes enables the application of stronger motion-planning frameworks such as trajectory optimization with Graphs of Convex Sets \cite{marcucci2023motion} and is currently a major roadblock in the adoption of these approaches.  
In this paper, we build upon IRIS-NP (Iterative Regional Inflation by Semidefinite \&
Nonlinear Programming) \cite{petersen2023growing} to significantly improve tunability, runtimes, and scaling to complex environments. IRIS-NP uses nonlinear programming paired with uniform random initialization to find configurations on the boundary of the free configuration space. Our key insight is that finding near-by configuration-space obstacles using sampling is inexpensive and greatly accelerates region generation. 
We propose two algorithms using such samples to either employ nonlinear programming more efficiently (\emph{IRIS-NP2}) or circumvent it altogether using a massively-parallel zero-order optimization strategy (\emph{IRIS-ZO}). 
We also propose a termination condition that controls the probability of exceeding a user-specified permissible fraction-in-collision, eliminating a significant source of tuning difficulty in IRIS-NP. We compare performance across eight robot environments, showing that IRIS-ZO achieves an order-of-magnitude speed advantage over IRIS-NP. IRIS-NP2, also significantly faster than IRIS-NP, builds larger polytopes using fewer hyperplanes, enabling faster downstream computation. Website: \href{https://sites.google.com/view/fastiris}{\texttt{https://sites.google.com/view/fastiris}}.

\keywords{collision-free motion planning, robot configuration space}
\end{abstract}

\section{Introduction}

A major challenge in robot motion planning is the need to simultaneously consider \emph{task space}, the world in which the robot physically resides, and \emph{configuration space}, $\calC$, the set of all possible robot configurations. The planner must produce a trajectory in the configuration space, but many constraints are formulated in the task space. Collision avoidance is particularly challenging, because even geometrically simple obstacles in task space can have  intractably complicated descriptions when transformed into configuration space through the robot's inverse kinematics. While there is work on constructing explicit configuration-space representations of workspace obstacles \cite{lozano1990spatial,bajaj1990generation,pan2015efficient}, these methods are intractable for the high degree-of-freedom (dof) robotic systems being used today.

The unavailability of obstacles' configuration-space descriptions has not prevented the development of a rich literature of motion planning algorithms.
Many approaches approximate the set of collision-free configurations ($\Cfree$) without explicitly constructing the individual obstacles.
Interval analysis and cell decompositions can 
approximate $\Cfree$ as the union of boxes~\cite{jaulin2001path}\cite[\S5-6]{latombe2012robot}, but these methods are computationally intractable for high-dimensional configuration spaces.
Perhaps the most widely used technique has been sampling-based planning, in which samples are drawn from $\Cfree$ and connected into a graph structure. 

These representations of $\Cfree$ are popular due to their simplicity and versatility.
However, they all struggle with the ``curse of dimensionality'' -- the memory use of the representation may grow exponentially with the dimension of the configuration space.
This has led the motion planning community to explore volumetric approximations of $\Cfree$ such as the union of spheres used by Yang and LaValle~\cite{yang2004sampling} or the polytopes constructed by Deits and Tedrake~\cite{deits2015computing}.
In contrast to the grid- or sampling-based approaches, the individual sets used in these approaches describe free-space regions rather than points.
Although these sets are harder to construct, they often enable a more concise approximation of $\Cfree$ and yield convex (probabilistic) collision-avoidance constraints.

Motion planning algorithms such as trajectory optimization with \gls*{gcs}~\cite{marcucci2023motion} leverage these representations to quickly produce high-quality, collision-free trajectories for high-dimensional robotic systems.
However, the performance of these planners is highly dependent on the properties of the convex sets. It is desirable for these sets to have large volumes in order to reduce the number of sets required to approximate $\Cfree$, while retaining simple descriptions to make downstream planning more efficient.
While creating perfectly collision-free sets is very costly, practical algorithms should provide a straightforward way to trade off between precision and runtime.

Recent results for planning with \gls{gcs} leverage the IRIS-NP algorithm~\cite{marcucci2023motion,jaitly2024paamp,cohn2023noneuclidean,cohn2023constrained}.
IRIS-NP takes in a collision-free \emph{seed} configuration and attempts to construct a convex, collision-free polytope containing it.
IRIS-NP, however, falls short of meeting the aforementioned criteria in practice. Its runtime is substantial and trading off between runtime and correctness (how much of the region is collision-free) proves challenging. IRIS-NP only terminates after failing to solve a user-specified number of nonlinear programs in succession, a time-consuming process dependent on this user-specified parameter acting as a proxy for correctness, when the actual relationship is unclear. Furthermore, these nonlinear programs must be run separately for every object in the scene, causing IRIS-NP to scale poorly with environment complexity.

In this paper, we improve upon IRIS-NP, focusing on a key subroutine that constructs hyperplanes to separate the seed point from obstacles.
Our improvements leverage the fact that we can evaluate thousands of configurations for collisions in the time it takes to solve a single nonlinear optimization problem.
We employ random sampling and collision checking to estimate the proportion of the polytope that is collision-free, providing a rigorous probabilistic certificate, and enabling intuitive tradeoffs between region correctness and computation times. We further present two algorithms that utilize sampling and collision checking to improve polytope generation: IRIS-ZO rapidly generates polytopes using a simple parallelized zero-order optimization strategy that requires no gradient computations and is easy to implement. IRIS-NP2 uses sampled collisions to seed nonlinear optimizations, increasing search success, and dramatically reducing the required number of programs. We demonstrate that both algorithms outperform IRIS-NP in terms of computation time and region quality.

\vspace{-0.2cm}
\section{Problem Formulation}\label{sec: preliminaries}
In this section, we introduce the problem formulation and discuss the required inputs and provided outputs of our proposed algorithms.

We aim to generate large convex polytopes in configuration space whose fraction in collision is less than a user-provided constant. More precisely, let $\lambda$ denote the Lebesgue measure in the free configuration space $\Cfree$. Given a user-specified admissible fraction in collision $\varepsilon\in(0,1)$ and confidence $\delta\in(0,1)$, we will compute positive-volume convex polytopes $\calR\subseteq\calC$ such that
\begin{gather}
    \Prob\left[\frac{\lambda(\calR\setminus\Cfree)}{\lambda(\calR)}>\varepsilon \right]\leq \delta.
    \label{eq:prob_polytope_cfree}
\end{gather}

Since obtaining a closed-form description of $\Cfree$ is intractable for general robotic systems \cite[\S4.3.3]{lavalle2006planning},\cite[\S 3]{latombe2012robot}, our algorithms utilize a task-space description. 
Such descriptions are readily provided via common robot description formats such as URDFs \cite{ros_urdf} or SDFs \cite{sdf_spec}.
%
%
We expect the robot to be described as $M$ sets  $\calG_i\subseteq\mathbb{R}^{N_\text{ts}}$ representing collision geometries in task space. Here, $N_\text{ts}\in \{2, 3\}$ is the dimension of the task space, and $q\in\calC$ is the configuration. The sets $\calG_i$ are represented in their body frame $B_i$. Using the monogram notation of \cite[\S 3.1]{manipulation}, each $\calG_i$ is paired with a configuration-dependent rigid-body transformation $^{W}X^{B_i}(q)\in SE(N_\text{ts})$ that defines the forward kinematics of $\calG_i$ in the world frame $W$. We use the shorthand $\calG_i(q)$ for $\calG_i$ expressed in world frame: 
\begin{equation*}
\mathcal{G}_i(q)=\{p_W\in\mathbb{R}^{N_{\mathrm{ts}}}\,|\,p_W=^{W}\!\!X^{B_i}(q)\cdot p_{B_i},\;p_{B_i}\in\mathcal{G}_i\}.
\end{equation*}
We let $\calG(q)= \{\calG_1(q),\calG_2(q), \dots, \calG_M(q)\}$ be the set of all collision geometries.

For a configuration $q$, the system is \emph{in collision} if there exists a \emph{valid}\footnote{Typically, only a subset of all collision pairs is considered for practical reasons.} pair of collision geometries $\calA(q), \calB(q) \in \calG(q)$ such that $\calA(q)\cap\calB(q)\neq\emptyset$. 
For a valid collision pair $\calA(q), \calB(q) \in \calG(q)$, the corresponding \emph{configuration-space obstacle} $\calO^{\calA\calB}(q)$ is implicitly defined as 
\begin{gather}
    \calO^{\calA\calB}(q) := \big\{q~|~\calA(q)\cap\calB(q)\neq\emptyset \big\}
\end{gather}
We avoid explicitly describing $\calO^{\calA\calB}$ by using nonconvex task-space constraints, as in  \cite{petersen2023growing}.

\section{Related Works}\label{sec:relworks}

Algorithms for computing positive-volume subsets of $\Cfree$ can be divided into two categories: those which require explicit descriptions of the obstacles in configuration space, and those that can use implicit descriptions.
Explicit descriptions of obstacles are generally only available for robots with simple kinematics (e.g. only prismatic joints).


If descriptions of all obstacles are given as convex sets, the original IRIS algorithm~\cite{deits2015computing} can construct large collision-free polytopes about a seed points using a series of convex optimizations. Such descriptions can be obtained from arbitrary meshes via approximate convex decomposition techniques~\cite{mamou2009simple}. Wu et. al.~\cite{wu2024optimal} use a similar approach to grow convex polytopes around an existing trajectory, towards producing a shorter, collision-free path.


Due to the complex kinematics of robotic manipulators, obstacles in configuration space are frequently given by implicit descriptions.
%
Yang and LaValle leveraged the kinematic Jacobian to relate motion in configuration-space and task space, allowing the construction of collision-free ellipsoids~\cite{yang2004sampling}.
Unfortunately, large numbers of ellipsoids are required to approximate even simple, low-dimensional configuration spaces.

The original IRIS algorithm has been extended to handle such implicit descriptions in two ways.
IRIS-NP uses nonlinear programming to find multiple locally separating hyperplanes to each obstacle, until the program becomes infeasible~\cite{petersen2023growing} -- an imprecise termination condition that often leaves some obstacle volume in regions.
Jaitly and Farzan modified IRIS-NP to use a nonuniform sampling strategy to seed the collision search program~\cite{jaitly2024paamp}.
The other option is to use a rational reparametrization of the kinematics to construct regions that are rigorously certified to be collision-free with sums-of-squares programming~\cite{dai2023certified}.
However, such optimizations are computationally expensive, and the regions are grown in a stereographic projection of configuration space, which distorts distances.

Alternatively, \cite{sarmientoy2005sample} directly decomposes three-dimensional spaces into polytopes only using sample-based collision-checking. Unfortunately, this approach requires dense sampling of the configuration space to produce large sets which is intractable in all but the simplest cases. 
\section{Background on IRIS Algorithms}\label{sec:background}
\tcdelete{The original IRIS algorithm \cite{deits2015computing} takes a set of $M$ convex obstacles $\mathcal{O}_1,\mathcal{O}_2, \dots, \mathcal{O}_M$ and produces a collision-free polytope in the same space as these obstacles. In this section we review the algorithm in order to set us up for understanding how to generalize the algorithm to nonconvex obstacles. }

\tcdelete{The algorithm takes as input a collision-free seed point $s\inR^n$, an initial ellipsoidal metric and the collection of obstacles $\mathcal{O}_{1:M}$ convex obstacles. The polytope is then computed iteratively by alternating between two steps, \textsc{SeparatingPlanes}, and \textsc{InscribedEllipsoid}, until a termination condition is met. This algorithm is reviewed in \S\ref{ssec:originaliris}. IRIS-NP \cite{petersen2023growing} generalizes the \textsc{SeparatingPlanes} step to grow polytopes in robot configuration space.
In this setting, the obstacles are nonconvex and defined implicitly in the robots task space. In \S\ref{ssec:irisincspace}, we review IRIS-NP, and provide a detailed formulation of the underlying nonlinear optimization. The new algorithms proposed in this paper are, in essence, new solution strategies for this this underlying nonlinear program.}

In this section, we review the key algorithms our approach builds upon.
The IRIS algorithm~\cite{deits2015computing} takes in a list of convex obstacles and produces a collision-free polytope around a seed point in the same space as these obstacles.
IRIS-NP~\cite{petersen2023growing} generalizes IRIS to operate in configuration space using nonlinear programming.

\subsection{IRIS with Convex Obstacles}\label{ssec:originaliris}

IRIS \cite{deits2015computing}, takes as input $N_{o}$ convex obstacles $\mathcal{O}_{i}$ and a collision-free seed point $s\in\Cfree$. IRIS then computes a large collision-free polytope by alternating between two convex optimizations: the \textsc{SeparatingPlanes} step, that produces a polytope conditioned on an initial ellipsoid, and the \textsc{InscribedEllipsoid} step, which updates the ellipsoid to the maximum-volume inscribed ellipsoid (MVIE) \cite[\S 8.4.2]{boyd2004convex} in the current polytope. These alternations guarantee the containment of the ellipsoid, and hence, guarantee that the volume of the MVIE is monotonically increasing across alternations. To understand the upcoming modifications to IRIS-NP, we review the \textsc{SeparatingPlanes} step in detail.

The \textsc{SeparatingPlanes} step starts with a collision-free ellipsoid $\calE$, 
\begin{align}\label{eqn:ellipsoidform}
    \calE = \{x~|~(x-c)^TE(x-c)\leq1, ~E\succ0\},
\end{align}
that is centered at $c$ with a symmetric, positive-definite matrix $E$. For each of the $N_o$ obstacles, IRIS computes a hyperplane that separates $c$ from $\calO_i$ and passes through the point in $\calO_i$ that lies closest to the ellipsoid center $c$ in the metric of the ellipsoid. This is done by solving
\begin{gather}\label{opt:uniformscaling}
    \quad\minz ~||x_i-c||_E^2~\subjectto~ x_i\in\calO_i,
\end{gather}
where $||x||_E^2 := x^TEx$. Given the optimum $x_i^\star$, the hyperplane
\begin{gather}
    \calH_i = \{x| a_i^Tx - b_i = 0\},\qquad
    a_i = E(x_i^\star-c),~
    b_i = a_i^Tx_i^\star
\end{gather}
separates $c$ from $\calO_i$ and does not intersect $\calE$~\cite[\S3.5]{deits2015computing}.
This step is performed for each obstacle. Finally, the intersection of the halfspaces $a_i^Tx\leq b_i$ yields an updated collision-free polytope. 


\subsection{Computing Separating Planes in Robot Configuration Space}\label{ssec:irisincspace}
\begin{figure}
    \centering\input{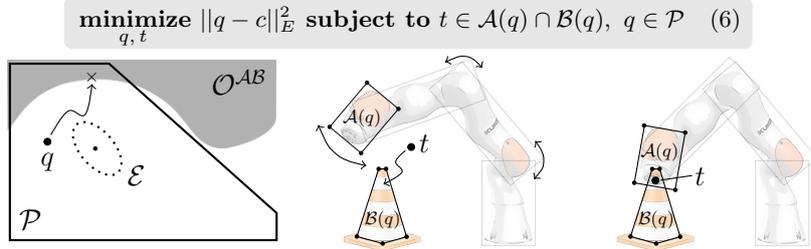}
    \caption{The nonlinear closest-point-in-obstacle program seeks the closest configuration $q$, under the ellipsoidal metric $\calE$, that lies inside of the current polytope $\calP$ and causes the collision pair $\calA,~ \calB$ to collide. Instead of constraining $q\in\mathcal{O}^{\mathcal{AB}}$ directly, we encode the collision constraint in task space with an auxiliary point $t$ that is constrained to lie in the in the intersection of both collision geometries. The picture on the right shows a feasible solution.  A locally optimal solution to the program is indicated by the cross in the cartoon on the left. }
    \label{fig:pointincollision program}
\end{figure}

IRIS requires convex obstacles to ensure the convexity of \eqref{opt:uniformscaling}.  However, configuration-space obstacles are generally non-convex.  IRIS-NP \cite{petersen2023growing} generalizes the closest-point-in-obstacle program (\ref{opt:uniformscaling}) for the configuration-space obstacle $\calO^{\calA\calB}$.
To avoid intractable descriptions of $\calO^{\calA\calB}$, the collision constraint is encoded  in task space. Optimization variables represent a point in each of the collision geometries, $t_\calA\in \calA$, $t_\calB\in \calB$, in their body frames, constrained by $^WX^{B_\calA}t_\calA = {^W}X^{B_\calB}t_\calB$, to coincide in task space when transformed through the forward kinematics. For clarity, in this paper, we write this as requiring a single point $t$ to lie in the intersection of the associated pair of collision geometries in task space. This is shown in \Cref{fig:pointincollision program}. In order to ensure that new hyperplanes address obstacles still relevant to the current polytope \calP, IRIS-NP requires the found points in collision to lie inside \calP. The full nonlinear program then reads:
\begin{subequations}
\newcommand{\alignspacing}{\;\;}
\label{opt:closestcollision}
\begin{align}
    \minz_{q,t} &\alignspacing ||q-c||^2_E\\
    \subjectto &\alignspacing t\in\calA(q)\cap\calB(q),\alignspacing q\in\calP.
\end{align}
\end{subequations}
Because configuration-space obstacles are generally non-convex, multiple hyperplanes may be necessary to separate $c$ from $\calO^{\calA\calB}$.

\begin{figure}
    \centering
    \input{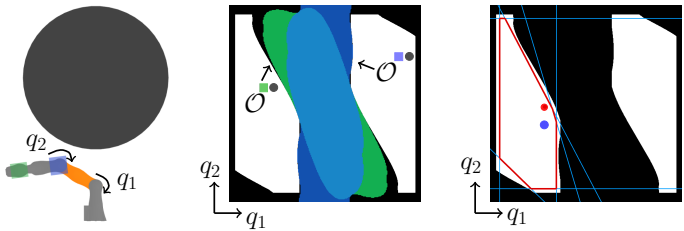}
    \vspace{-0.3cm}
    \caption{A visualization of an IRIS-NP region for a simple two dof system. \textit{Left}: Two dof robot arm with a disk shaped obstacle. Two selected collision geometries of the system are highlighted in green and blue. \textit{Center}: Configuration space of the system. The black regions correspond to collisions. Two of the configuration-space obstacles are highlighted in blue and green, corresponding to the configurations where the blue and green collision geometries intersect the disk obstacle. \textit{Right}: Resulting region (red outline), when seeding at the red dot. The configuration in left frame is shown by the blue dot.}
    \label{fig:2dofsys}
\end{figure}

To construct the separating hyperplanes for the collision pair $(\calA,\calB)$, IRIS-NP repeatedly finds locally optimal solutions $q^{\star}$ to (\ref{opt:closestcollision}), adding the hyperplane
\begin{gather}
    \calH_i = \{x| a_i^Tq^\star - b_i + \Delta= 0\},\qquad
    a_i = E(q^\star-c),~
    b_i = a_i^Tq^\star,
\end{gather} 
to $\calP$ for each of the found points $q^{\star}$ in collision. More precisely, the polytope $\calP$ is updated by intersecting $\calP$ with the halfspace $\{q|a_i^Tq\leq b_i-\Delta\}$, where $\Delta>0$ is a user-specified stepback. This stepback ensures that the collision configuration $q^\star$ is excluded from \calP, and prevents the need for an infinite number of hyperplanes to exclude a non-convex obstacle. This is repeated until a fixed number of attempted solves of (\ref{opt:closestcollision}) fail; IRIS-NP interprets this failure as a suggestion that $\calP$ is sufficiently collision-free. However, note that due to the local nature of nonlinear programming, a single failed solve often provides inconclusive information about the feasibility of the overall program.

To summarize, the IRIS-NP strategy to approximating \textsc{SeparatingPlanes} in robot configuration space is to create a list of all valid collision pairs ordered by their task-space distance, and, for each collision pair, repeatedly find locally optimal solutions to (\ref{opt:closestcollision}) and update $\calP$ until a user-specified number of consecutive solve attempts fail. Through this procedure, IRIS-NP strives to make \calP\ collision-free with respect to all valid collision pairs. See \Cref{fig:2dofsys} for an illustration.

\section{Improving the Separating Hyperplanes Routine}\label{sec:separating_hyperplanes}
This section discusses how we leverage sampling to produce a rigorous termination condition for \textsc{SeparatingPlanes} in robot configuration space (see \cref{ssec:irisincspace}), and two improved approaches for solving the step, yielding the new algorithms IRIS-ZO and IRIS-NP2, whose parameters are given in \Cref{tab:params}. Both of these new formulations follow the same alternation scheme as IRIS-NP, given in \Cref{alg:Iristemplate}.
In particular, we still initialize the ellipsoid with a small ball of radius $r_\text{start}$ and employ the same overall termination conditions for the alternations as \cite[\S II.D]{petersen2023growing}, such as the seed point $s$ no longer being contained in $\calP$, reaching a maximum number of alternations, or achieving convergence of the ellipsoid volume. Our proposed termination condition for \textsc{SeparatingPlanes} decides when $\calP$ meets a correctness criterion based on polytope fraction in collision, not when to terminate the alternations. 
We discuss the proposed termination condition in \Cref{ssec:unadaptivetest}, before discussing IRIS-ZO in \Cref{ssec:fastiris}, and IRIS-NP2 in \Cref{ssec:initiris}.
\begin{table}
    \centering
    \begin{tabular}{cccc}
      Parameter & Description & IRIS-ZO & IRIS-NP2  \\
        \hline
        $\varepsilon$ & Admissible fraction of the region in collision & \cmark&\cmark  \\
        $\delta$& Max admissible uncertainty & \cmark & \cmark \\
        $\Delta$ & Configuration margin, i.e.``step back'' & \cmark & \cmark \\
        $T$& Termination condition (as in \cite{petersen2023growing}) & \cmark & \cmark\\
        $N_p$ & Number of optimized particles per inner iteration & \cmark &\cmark\\
        $N_b$ & Number of bisection steps & \cmark &\xmark\\
        $N_f$ & Max number of hyperplanes added per inner iteration & \cmark& \cmark\\
   \end{tabular}
    \caption{Glossary of algorithm parameters.}
    \label{tab:params}
\end{table}
\begin{algorithm}
\SetAlgoLined
\caption{Template for the new IRIS algorithms}
 \label{alg:Iristemplate}
\SetKwInput{Input}{Input}
\SetKwInput{Output}{Output}
\SetKw{KWAlgorithm}{Algorithm:} 
\Input{
Domain $\calD\subseteq\calC$, collision-free seed $s\in\calD$, options $O$.
}
\Output{
Polytope $\calP\subseteq\calD$
}

\KWAlgorithm{}
$\calE \gets \textsc{Ball}(s, r_\text{start})$, $i\gets 1$

\While{not done}{ 
$\calP \gets \textsc{SeparatingPlanes}(\calD, \calE, i, O)$ \Comment{\textbf{employs new term. cond.}}\\
$\calE \gets \textsc{InscribedEllipsoid}(\calP)$\\
$i\gets i+1$
}
\Return $\calP$
\end{algorithm}

\subsection{Termination Condition for the Separating Planes Step}\label{ssec:unadaptivetest}
In this section, we discuss a termination condition that allows a user to specify a desired bound on the fraction of the volume of the  polytope \calP\ in collision. In the following, let $\varepsilon_{tr}:= \lambda(\calP\setminus\Cfree)/\lambda(\calP)$ denote the true fraction in collision of \calP, where $\lambda$ is the Lebesgue measure over $\Cfree$, and $\varepsilon$ is the specified admissible fraction in collision.  
A naive approach may be to sample uniformly in the polytope \calP, estimate the fraction of \calP\ in collision $\hat\varepsilon$ to be equal to the fraction of these samples that are in collision, and terminate if $\hat\varepsilon\leq \varepsilon$.  However, each time we perform this check, there is some chance of underestimating $\varepsilon_{tr}$ such that we incorrectly terminate with a polytope with $\varepsilon_{tr} > \varepsilon$. The probability of false termination accumulates over the multiple evaluations of this condition.   

Instead, we propose a statistical test that controls the probability of falsely claiming a polytope is sufficiently collision-free and terminating. 
For some user-specified uncertainty $\delta$, 
a correct termination condition allows a region with $\varepsilon_{tr}>\varepsilon$ to be returned with probability at most $\delta$.
To accomplish this, we pair union bounds with a simple statistical test based on a Chernoff bound.

We sample a batch of $M$ points $q_i\in\calP$ uniformly. We then assign $X_i = 0$ if $q_i\in\Cfree$ and $X_i=1$ otherwise. Note that $X_i\sim \mathrm{Bernoulli}(\varepsilon_{tr})$. The employed bound reads as follows.

\begin{theorem}[Sample Bound]\label{thm1}
    Let $\varepsilon_{tr}\geq \varepsilon\geq 0$, $(X_i)_{M\geq i\geq1}\sim\mathrm{Bernoulli}(\varepsilon_{tr})$, with $M \geq 2 \log(1/\delta)/(\varepsilon \tau^2)$, for any fixed parameter $\tau>0$, and $\delta\in(0,1]$. Define $\bar X_M:=\frac{1}{M}\sum_{i=1}^MX_i$. It holds that
    \begin{align}
        \Prob \left[\bar X_M  \le (1-\tau) \varepsilon \right] \le\delta.
    \end{align}
\end{theorem}

\noindent\textbf{Proof} Observe that $\Prob\left[\bar X_M  \le (1-\tau) \varepsilon \right] \le \Prob\left[\bar X_M  \le (1-\tau) \varepsilon_{tr} \right]$. If $\tau>1$ both sides evaluate to 0, and otherwise $(1-\tau) \varepsilon \leq (1-\tau) \varepsilon_{tr}$. Next, we use the standard multiplicative Chernoff bound, $\Prob\left[\bar X_M  \le (1-\tau) \varepsilon_{tr} \right]\leq e^{-M\varepsilon_{tr}\tau^2/2},$ 
e.g. given in \cite[\S A, Thm A.1.15]{alon2016probabilistic}, and use the fact that $e^{-M\varepsilon_{tr}\tau^2/2} \le  e^{-M\varepsilon\tau^2/2}$. We have now shown that $\Prob\left[\bar X_M  \le (1-\tau) \varepsilon_{tr} \right]\leq e^{-M\varepsilon\tau^2/2}$, and the final result follows from evaluating $M = 2 \log(1/\delta)/(\varepsilon \tau^2)$.\qed

\begin{definition}[Unadaptive Test]\label{def:unadaptive_test} 
Let $\mathsf{UnadaptiveTest}(\delta,\varepsilon,\tau)$ collect $M = \lceil2 \log(1/\delta)/\varepsilon \tau^2\rceil$ samples $(X_i)_{1 \le i \le M}$. $\mathsf{UnadaptiveTest}$ returns \texttt{reject} if $\bar{X}_M > (1-\tau)\varepsilon$, and \texttt{accept} otherwise.
\end{definition}
\begin{corollary}[Controlling False Accept] \label{corr1}
Suppose $\varepsilon_{tr} \ge \varepsilon$, i.e. the true fraction in collision is higher than allowed,  and we have $(X_i)_{1 \le i \le M} \sim \mathrm{Bernoulli}(\varepsilon_{tr})$. Then,  $\mathsf{UnadaptiveTest}(\delta,\varepsilon,\tau)$ rejects with probability at least $1 - \delta$. 
\end{corollary}

Thus, our test behaves as follows: Assume we are given a polytope $\calP$, an admissible fraction in collision $\varepsilon$, an admissible target uncertainty $\delta$, and a batch of $M$ uniform samples in $\calP$ with $M = \lceil2 \log(1/\delta)/(\varepsilon \tau^2)\rceil$. If it holds that $\calP$ does \textit{not} meet the specifications, i.e. $\varepsilon_{tr}\geq\varepsilon$, then $\mathsf{UnadaptiveTest}(\delta,\varepsilon,\tau)$ rejects $\calP$ with probability greater than $1-\delta$; the test only falsely accepts $\calP$ with probability less than $\delta$. Observe that if $\tau$ decreases, we require many samples and the test becomes expensive. Conversely, if we increase $\tau$, the number of samples decreases, but the test starts rejecting every $\calP$. In practice, we choose $\tau = 0.5$ to balance power and computational cost of the test.

Within \textsc{SeparatingPlanes}, for both algorithms, hyperplanes are iteratively added to $\calP$ until sufficient separation from the configuration-space obstacles is achieved. We refer to these iterations as \emph{inner iterations}. In order to decide whether the polytope is sufficiently collision-free at the $k$-th inner iteration, we run $\mathsf{UnadaptiveTest}(\delta_k,\varepsilon,\tau)$. We can bound the probability of \textsc{SeparatingPlanes} terminating falsely after $K$ iterations by the union bound

\begin{gather}
    \Prob(\text{``false accept'' if }\lambda(\calP\setminus\Cfree)/\lambda(\calP)>\varepsilon)\leq \sum_{k=1}^K\delta_{k}.
\end{gather}

In order to control the probability of falsely accepting, i.e. terminating, after an unknown number of inner iterations, in the case where a single outer iteration is to be run, we simply select $\delta_k$ according to the sequence  $\delta_k = 6\delta/(\pi^2k^2)$ for the $k$-th test, as this sequence sums to $\delta$ as $k\rightarrow\infty$. 
If an unknown number of \emph{outer iterations} (i.e. alternations between updating the polytope and the ellipsoid) are to be run, we must account for the additional \textsc{UnadaptiveTest} queries.
In order for the probability of falsely accepting to sum to $\delta$ over infinite inner and outer iterations, we run $\mathsf{UnadaptiveTest}(\delta_{i,k},\varepsilon,\tau)$, in the $i$-th outer iteration,  after the $k$-th inner iteration, with $\delta_{i,k} := 36\delta/(\pi^4i^2k^2)$.

Note that this termination condition assumes uniform samples in $\calP$. In practice, we generate approximately uniform samples using hit-and-run sampling \cite{lovasz1999hit} with a sufficiently high number of mixing steps.

\subsection{The IRIS-ZO Algorithm}\label{ssec:fastiris}
\begin{figure}
    \centering
    \includegraphics[width = 0.95\linewidth]{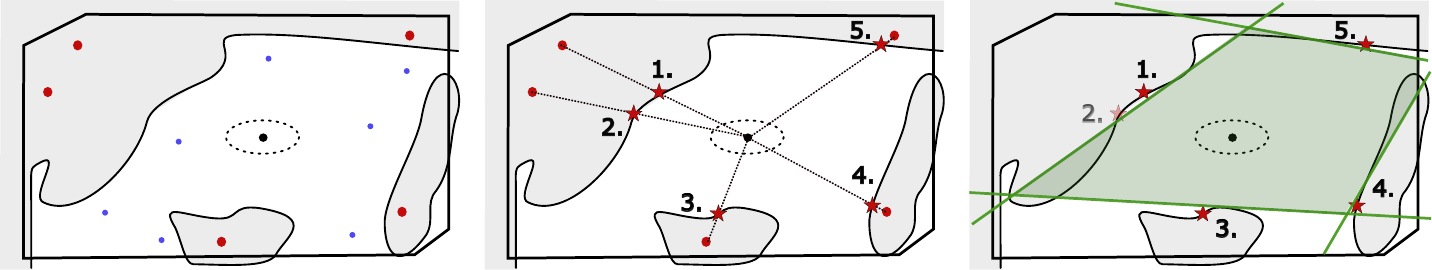}
    \vspace{-0.1cm}
    \caption{ One iteration of \textsc{ZeroOrderSeparatingPlanes}. \textit{Left}: Configuration space with grey configuration-space obstacles. First, a batch of samples is drawn uniformly in the current polytope and checked for collisions. \textit{Center}: The found collisions $\calS_\text{col}$, shown in red, are brought closer to the center of the ellipsoid using bisection. The updated candidate points $\calS_\text{col}^\star$(red stars), are sorted by their weighted squared distance to the center of the ellipsoid under the ellipsoidal metric. \textit{Right} - Hyperplanes are iteratively added for the non-redundant candidates in order. Note that the second candidate is made redundant by the first hyperplane, and therefore only four hyperplanes are added to $\calP$.}
    \label{fig:fastiris}
\end{figure}

The IRIS-ZO algorithm uses a simple parallelized zero-order optimization strategy to directly solve \textsc{SeparatingPlanes} in \Cref{alg:Iristemplate} for all collision pairs simultaneously. We call this subroutine \textsc{ZeroOrderSeparatingPlanes} and summarize it in \Cref{alg:fastseparating}.  \Cref{fig:fastiris} illustrates how \textsc{ZeroOrderSeparatingPlanes} optimizes hyperplanes.
\begin{algorithm}
\SetAlgoLined
\caption{\textsc{ZeroOrderSeparatingPlanes}}
\label{alg:fastseparating}
\SetKwInput{Input}{Input}
\SetKwInput{Output}{Output}
\SetKw{KWAlgorithm}{Algorithm:} 
\Input{
Domain $\calD\subseteq\calC$, ellipsoid $\calE=(E,c)$, current outer iteration $i\in\mathbb N$.
}
\Output{
Polytope $\calP\subseteq\calD$ satisfying \eqref{eq:prob_polytope_cfree} for $(\varepsilon,\delta_i)$.
}

\KWAlgorithm{}
$k \gets 0$, $\calP\gets\calD$\\
\While{\textsc{True}
}{
    $\calS \sim \textsc{UniformSample}(\calP, M)$\\
    $\calS_\text{col} \gets \textsc{PointsInCollision}(\calS)$\\
    \textbf{If} $\textsc{UnadaptiveTest}( \delta_{i,k}, \varepsilon, \tau)$ returns \texttt{accept} \textbf{then} $break$.\\
    $\calS_\text{col}^\star \gets \textsc{UpdatePointsViaBisection}(\calS_\text{col}, c)$\\
    $\calP \gets \textsc{OrderAndPlaceNonRedudnantHyperplanes}(\calP, \calE, \calS_\text{col}^\star, \Delta)$\\
    $k \gets k + 1$
}
\Return $\calP$
\end{algorithm}
The \textsc{ZeroOrderSeparatingPlanes} step constructs a probabilistically collision-free polytope $\calP$ by repeating 5 steps until the statistical test passes. We take as input the domain $\calD$, the current ellipsoid $\calE$, the current outer iteration $i$ and options that are summarized in \Cref{tab:params}. Here, we assume that the center of the current ellipsoid is collision-free. We then initialize $\calP$ with the domain $\calD =\{q|A_\calD q\leq b_\calD\}$ (typically a polytope describing the joint limits). \looseness=-1

The first step is to uniformly sample a batch of configurations $\calS$ in $\calP$ via hit-and-run sampling \cite{lovasz1999hit} with a batch size of $\text{max}\{M, N_p\}$. Then, we find all configurations $\calS_\text{col}\subseteq\calS$ that are in collision by running a collision checker. These samples are used to check the termination condition by running the unadaptive test, returning $\calP$ if it accepts. More precisely, counting the collisions $|\calS_\text{col}^{M}|$ in the first $M$ samples in $\calS$ and accepting if $|\calS_\text{col}^{M}|\leq M(1-\tau)\varepsilon$.

If the probabilistic test fails, we then add hyperplanes to the polytope. We produce candidate solutions for the closest point in collision program \eqref{opt:closestcollision} based on the first $N_\text p$ configurations $q$ in $\calS_\text{col}$. For each $q$, we use bisection search to step toward the center $c$ of the ellipsoid, while ensuring that $q$ is still in collision. In practice, we bisect a fixed number $N_b$ of times and pick the configuration $q^\star$ that is closest to $c$ and still in collision. This yields our updated batch $\calS^\star$. This strategy is guaranteed to converge to the boundary of an obstacle for each configuration $q$, but not necessarily the obstacle containing the original $q$ or the one that is closest to $c$. Because this optimization procedure is gradient-free and only relies on a collision checker, it is highly parallel and its performance scales with available computational resources. 
Finally, we order all candidates $q^\star\in\calS^\star_\text{col}$ by  ellipsoidal metric in ascending order.
We repeatedly pick the closest candidate $q^\star$ in the polytope and add a hyperplane at that point tangent to the ellipsoid, until we hit $N_f$ non-redundant hyperplanes or run out of candidates.

\subsection{The IRIS-NP2 Algorithm}\label{ssec:initiris}
\vspace{-0.1cm}
\begin{figure}
    \centering
    \vspace{-0.2cm}\includegraphics[width = 0.9\linewidth]{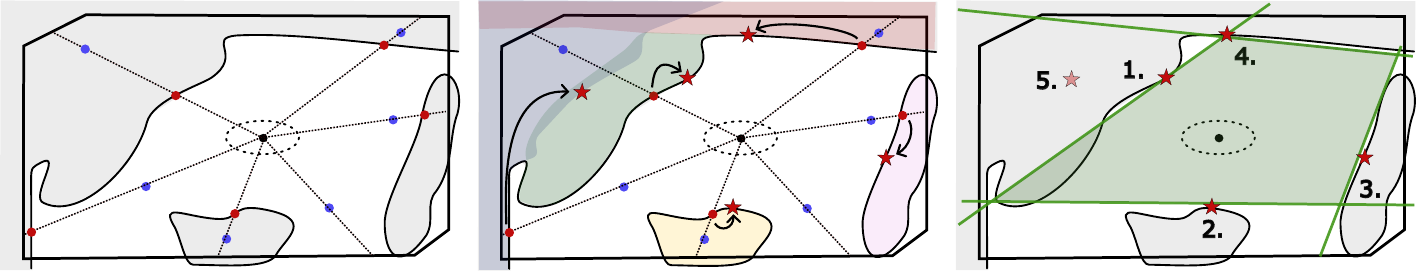}
    \vspace{-0.2cm}
    \caption{Execution of \textsc{NP2-SeparatingPlanes} using the ray collision finder for \textsc{GetConfigInCollision}.  Blue dots represent uniform samples in the polytope.  Red dots represent the results of the corresponding discrete line searches.  Stars represent the results of local nonlinear searches (\ref{opt:closestcollision}).  On the right, the corresponding hyperplanes (green) form the updated polytope.}
    \label{fig:ray-iris}
    \vspace{-0.1cm}
\end{figure}

The IRIS-NP2 algorithm (\Cref{alg:initseparating}) places hyperplanes more carefully than IRIS-ZO, increasing runtime in favor of larger regions with fewer hyperplanes, while still improving upon IRIS-NP runtimes. Like IRIS-NP, IRIS-NP2 uses nonlinear programming, solving \eqref{opt:closestcollision} to find separating hyperplanes.  In contrast to IRIS-NP, IRIS-NP2 always initializes NLP searches with feasible initial guesses $q_0$, prioritizing initial guesses with lower ellipsoid metric $||q_0-c||^2_E$.  These initial guesses, along with the corresponding pair of bodies in collision, $(\mathcal{A}, \mathcal{B})$ such that $\mathcal{A}(q)\cap \mathcal{B}(q)\neq \emptyset$, are obtained via a subroutine \textsc{GetConfigInCollision}, two options for which are outlined at the end of this section.
\begin{algorithm}
\SetAlgoLined
\caption{\textsc{NP2-SeparatingPlanes}}
\label{alg:initseparating}
\SetKwInput{Input}{Input}
\SetKwInput{Output}{Output}
\SetKw{KWAlgorithm}{Algorithm:} 
\Input{
Domain $\calD\subseteq\calC$, ellipsoid $\calE=(E,c)$, current outer iteration $i\in\mathbb N$
}
\Output{
Polytope $\calP\subseteq\calD$ satisfying \eqref{eq:prob_polytope_cfree} for $(\varepsilon,\delta_i)$.
}

\KWAlgorithm{}
$k \gets 0$, $\calP\gets\calD$\\
\Repeat{
    $\mathsf{UnadaptiveTest}(\delta_{i,k}, \varepsilon, \tau)$ \text{ returns \texttt{accept}}
}{$i\leftarrow 0$, $n_f\leftarrow 0$\\
    \While{$i < N_p$ and $n_f < N_f$}{
        $q_0,(\mathcal{A},\mathcal{B})\gets$\textsc{GetConfigInCollision}\\
        \If{$q_0$ is not \texttt{None}}{$q^*\gets$ solve (\ref{opt:closestcollision}) with initial guess $q_0$, with pair $(\mathcal{A},\mathcal{B})$\\
        $a = \frac{E (q^*-c)}{||E (q^*-c)||}$, $b = a^\top q^*-\Delta$\\
        Add hyperplane defined by $(a, b)$ to \calP\\
        $n_f\leftarrow n_f + 1$}
        $i \leftarrow i + 1$
    }
    $k\gets k+1$\\
}
\Return $\calP$

\end{algorithm}

This strategy of always using a feasible initial guess presents a major advantage over IRIS-NP, as IRIS-NP spends substantial time in NLP solves that return infeasible.
In part, this property of IRIS-NP is due to initializing NLP searches with uniform samples $q_0$, in the hopes that the nonlinear program solver can pull those $q_0$ not in collision into collision to return feasible optimized solutions $q^\star$.
In practice, this frequently fails even when the polytope does contain collisions. Furthermore, as discussed in \Cref{ssec:irisincspace}, IRIS-NP \textit{requires} that many consecutive NLP solves return infeasible in order to terminate. Assessing termination readiness via the Bernoulli trials is faster, allowing the NLP, which is slow to solve, to be used only to construct hyperplanes.
Another advantage of this strategy is it scales better with environment complexity. IRIS-NP must solve a minimum number of NLPs for every valid collision pair. This quantity grows quickly with scene complexity, while many of these pairs never actually collide, resulting in infeasible NLPs and wasted computation time. IRIS-NP2 only considers pairs known to collide, and hence solves only feasible NLPs.
%
%
We present two options for \textsc{GetConfigInCollision}, with differing costs and benefits.

For the \emph{Greedy Collision Finder}, all samples from the Bernoulli trial that are in collision are aggregated and sorted in ascending order of distances to the center point of the ellipsoid, w.r.t. the ellipsoidal metric.
(In effect, $N_p$ is dynamically set to equal the number of collision particles.)
When queried, this subroutine returns the sample with the next-lowest ellipsoid metric, until all in-collision samples from the Bernoulli trial have been exhausted.
If a sample is not in the polytope (due to new hyperplanes from an earlier sample), it would no longer be a feasible initial guess for \Cref{opt:closestcollision}, so \texttt{None} is returned.

The \emph{Ray Collision Finder} (\Cref{fig:ray-iris}) prioritizes the quality of the constructed hyperplanes via an inexpensive line search to find configurations in collision near the ellipsoid center.
The ray collision finder takes a subset of the samples drawn for the Bernoulli trial, and, for each sample, steps outward in discrete steps along the ray from the ellipsoid center through the sample.  The first step in collision is returned.  If a step has exited the polytope, \texttt{None} is returned. This discrete search serves a purpose similar to the objective of (\ref{opt:closestcollision}): to limit the hyperplanes needed, we desire to add hyperplanes at the closest (in the ellipsoid metric) points in collision.  Because the discrete search is not restricted to a specific collision pair, it may find configuration-space obstacles closer than \eqref{opt:closestcollision}.  In contrast to the bisection search performed by IRIS-ZO, this line search sweeps monotonically outward from the ellipsoid center in an attempt to find the closest collision; the IRIS-ZO bisection search steps inward from the sample, finding a point on a boundary of a configuration-space obstacle. (The ray may pass through several such boundaries.)
\vspace{-0.2cm}
\section{Experiments}\label{sec:experiments}
\vspace{-0.1cm}

\begin{figure}[h]
\centering
    \begin{tikzpicture}
        \matrix (m) [anchor=north, yshift=-30.1cm, column sep=2pt, row sep=-4pt] {
            \node {\includegraphics[width=0.20\linewidth]{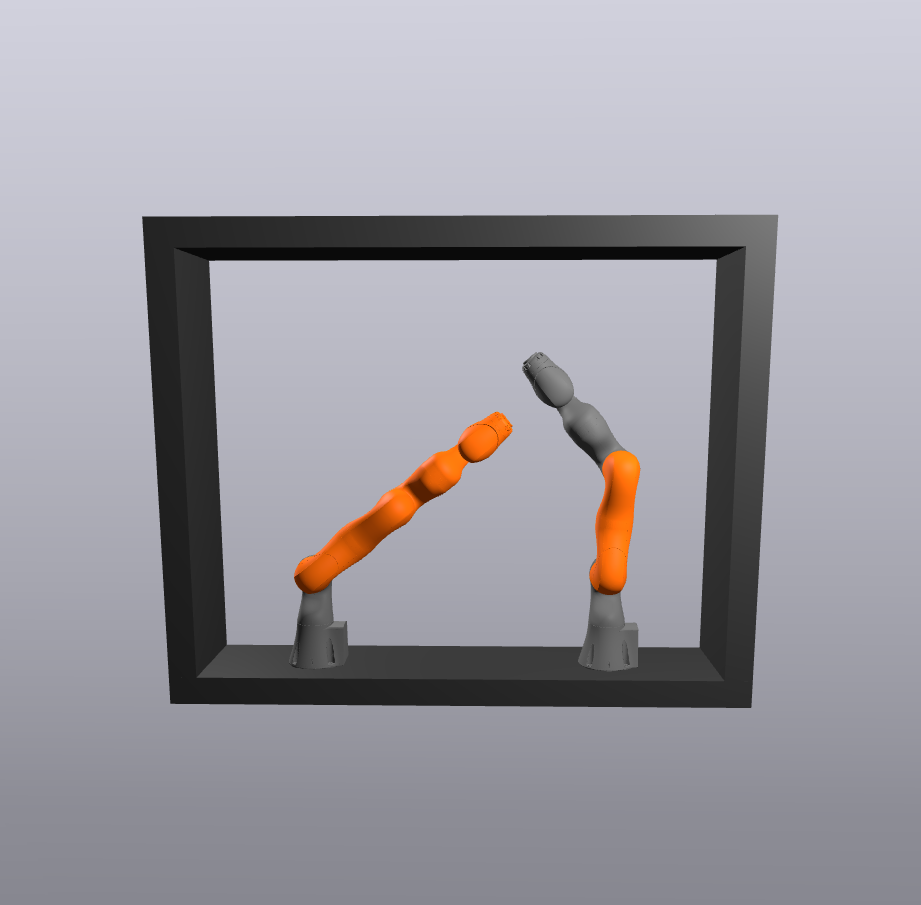}}; \node at (-0.1, -2.55) {\texttt{1. Flipper}};&
            \node {\includegraphics[width=0.20\linewidth]{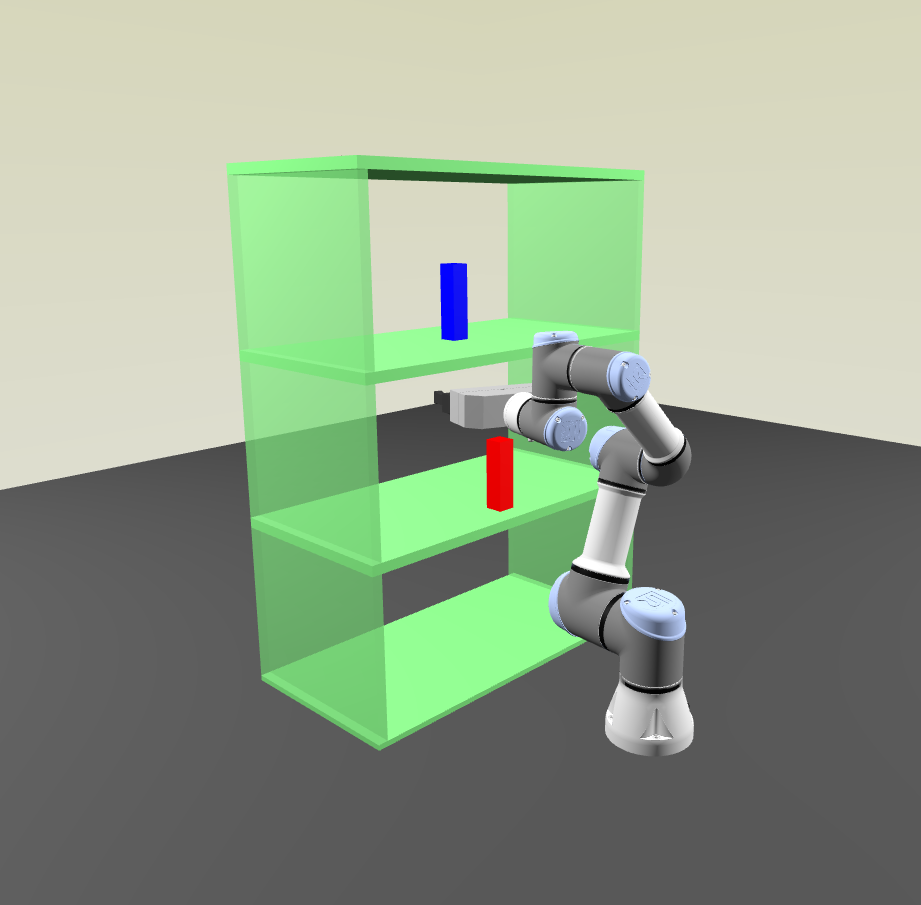}};\node at (-0.1,  -2.55) {\texttt{2. UR3}}; &
            \node {\includegraphics[width=0.20\linewidth]{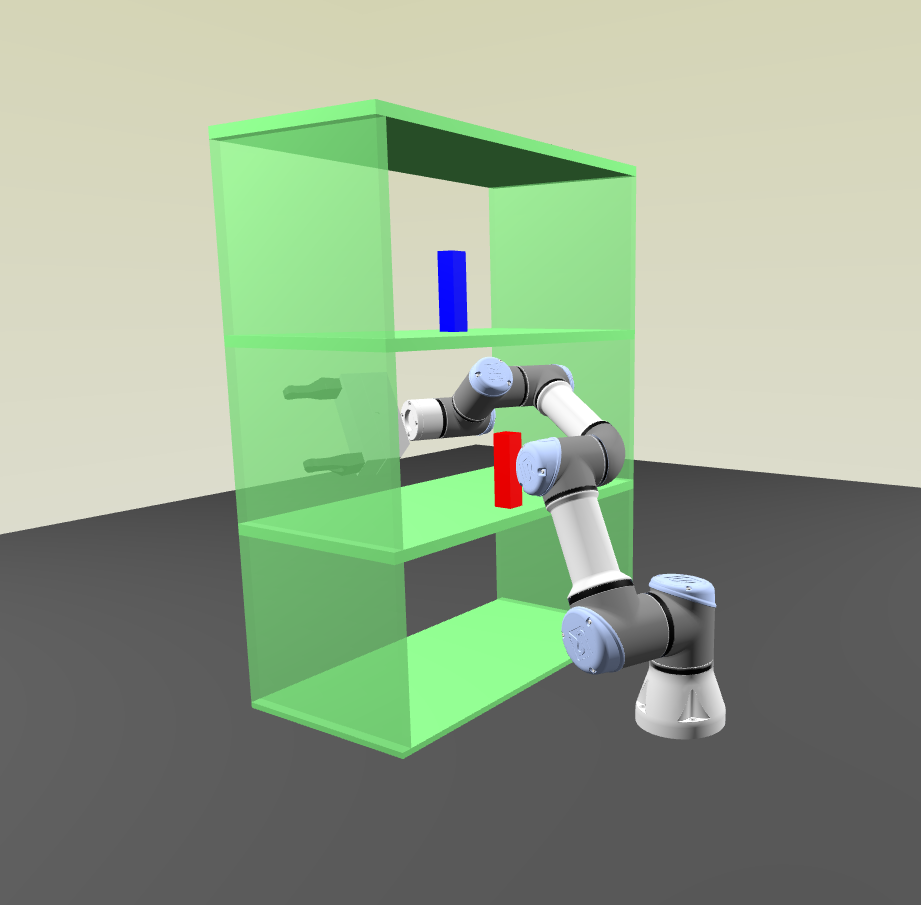}}; \node at (-0.1,  -2.55) {\texttt{3. UR3Wrist}};&
            \node {\includegraphics[width=0.20\linewidth]{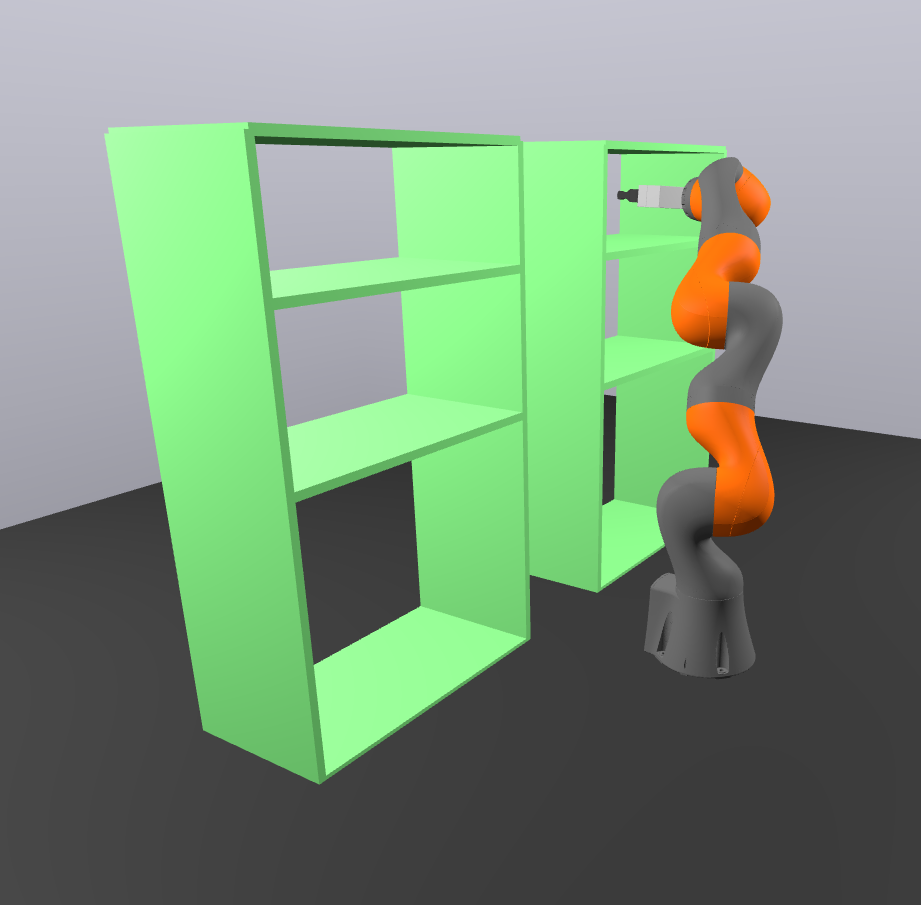}}; \node at (-0.05,  -2.55) {\texttt{4. IIWAShelf}};\\
            \node {\includegraphics[width=0.20\linewidth]{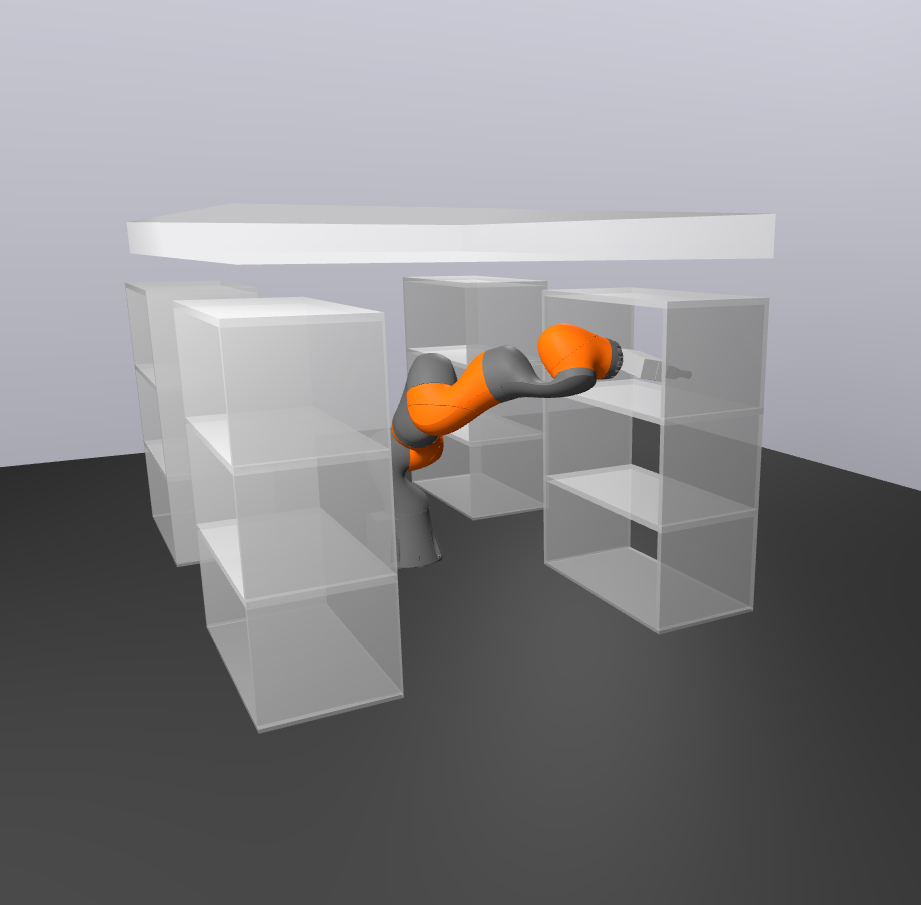}}; \node at (-0.1,  -2.55) {\texttt{5. 4Shelves}};&
            \node {\includegraphics[width=0.20\linewidth]{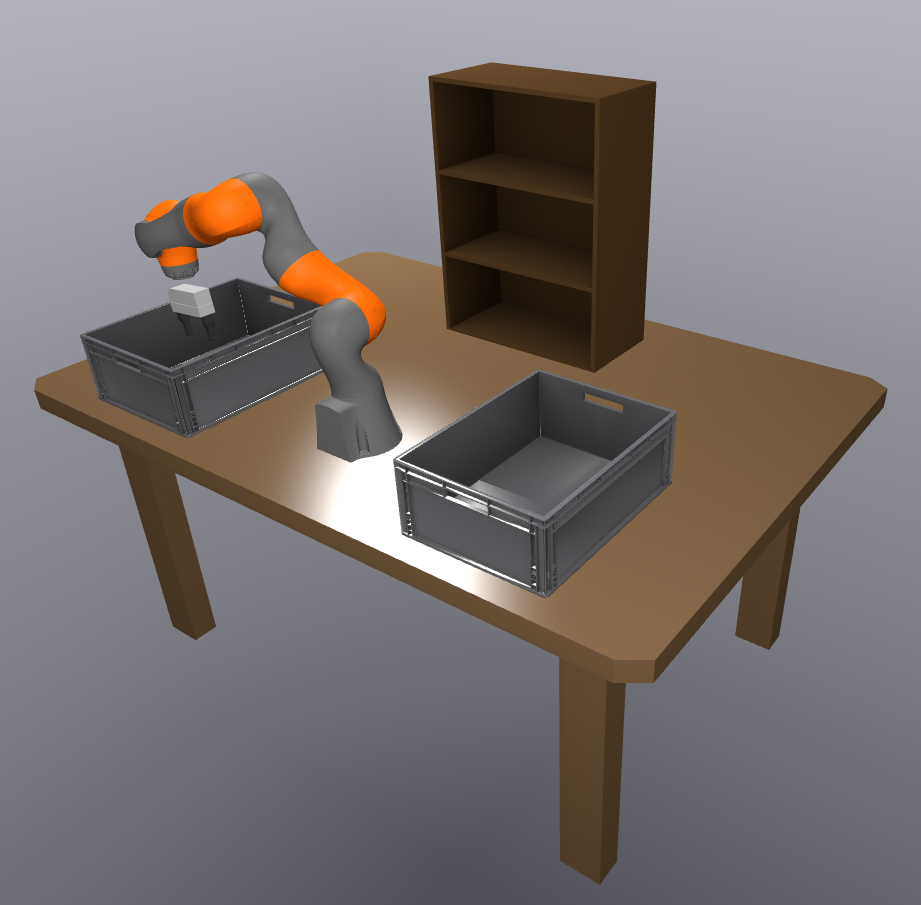}}; \node at (-0.1,  -2.55) {\texttt{6. IIWABins}};&
            \node {\includegraphics[width=0.20\linewidth]{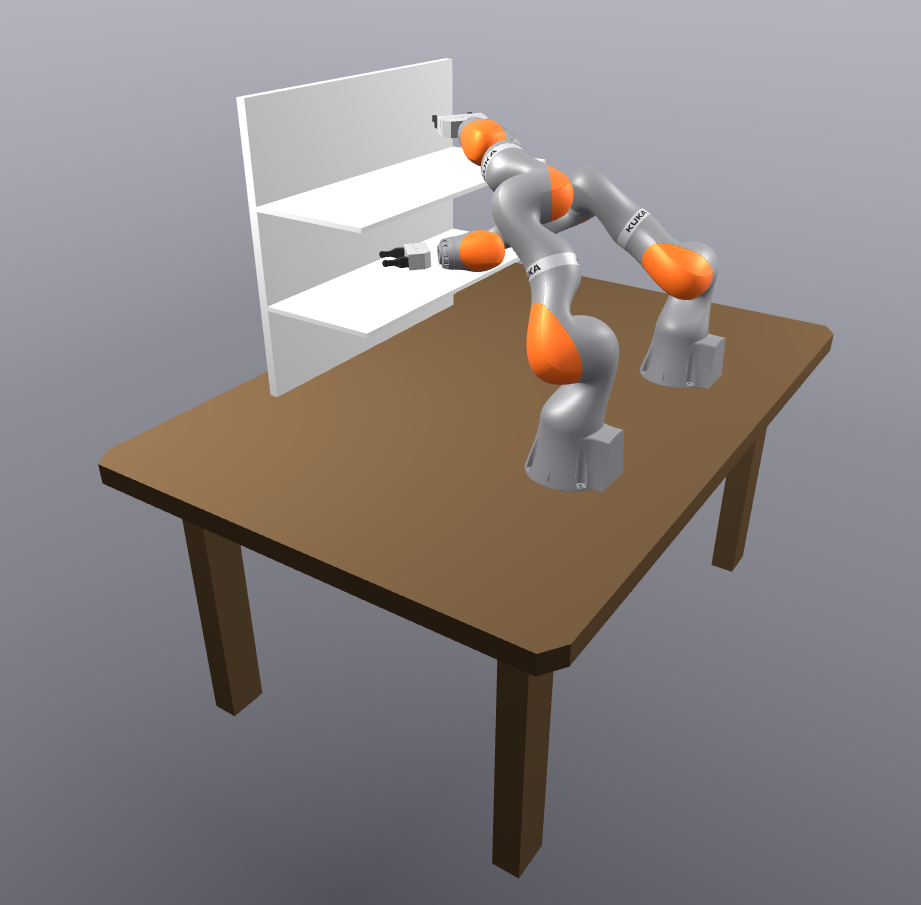}}; \node at (-0.1,  -2.55) {\texttt{7. 2IIWAs}};&
            \node {\includegraphics[width=0.20\linewidth]{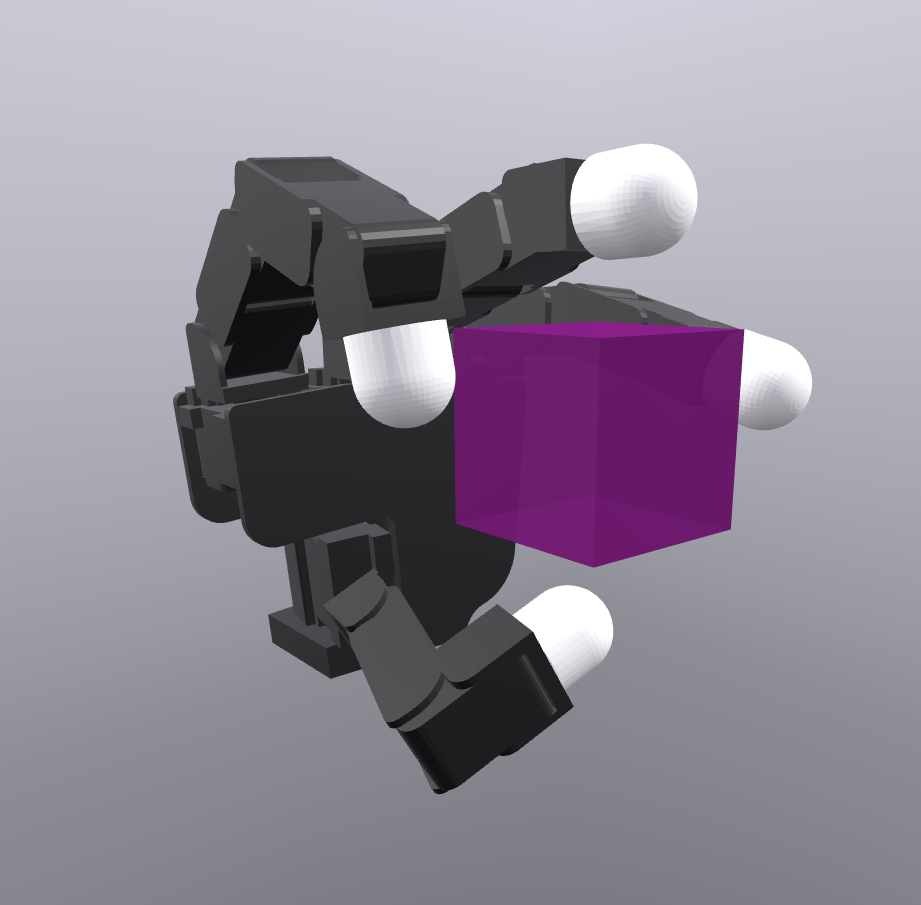}};\node at (-0.1, -2.55) {\texttt{8. Allegro}}; \\
        };
        \node[anchor=north, yshift=0.25cm] at (m.south) {
        \begin{minipage}{0.95\linewidth}
        \resizebox{\linewidth}{!}{
            
            \begin{tabular}{lcccccccc}
                \toprule
                Domain&
                  \texttt{Flipper}&\texttt{UR3}&\texttt{UR3Wrist} & \texttt{IIWAShelf}& \texttt{4Shelves}& \texttt{IIWABins} & \texttt{2IIWAs}&\texttt{Allegro}\\
                  \midrule
                  \# dof &3&5&6&7&7&7&14&15\\
                  \# col. pairs &137&142&142&172&302&212&264&187\\
                  $\text{vol}(\Cfree)/ \text{vol}(\calC)$& 0.149& 0.268& 0.254& 0.480& 0.117& 0.577& 0.536& 0.261\\
                \bottomrule
            \end{tabular}
            }
        \end{minipage}
        };
    \end{tikzpicture}
    \vspace{-0.2cm}
    \caption{The systems we use to benchmark our algorithms, along with the number of degrees of freedom, number of collision pairs, and the ratio of the volume of the free configuration space relative to the entire configuration space $\calC$. The ratio $\text{vol}(\Cfree)/ \text{vol}(\calC)$ is estimated with $10^7$ samples uniformly drawn from $\calC$.
    }
    \label{fig:environments}
\end{figure}

To evaluate our proposed changes to the IRIS-NP algorithm, we designed a comprehensive benchmark. This benchmark includes eight different robotic systems, described in \Cref{fig:environments}. For each benchmark, we select 10 seed configurations manually, placing the robot in a variety of configurations in its environment. Each experiment is run for a single outer iteration, to isolate performance differences in the \textsc{SeparatingPlanes} steps.

We compare the approaches for two different settings: a ``fast'' setting, where we request the region to be $90\%$ collision-free, with $90\%$ confidence, and a ``precise'' setting, where we request the region to be $99\%$ collision-free with $95\%$ confidence. IRIS-NP cannot take these arguments directly, so we hand-tune the required number of consecutive infeasible solves to roughly match the average collision fractions achieved by our algorithms. For IRIS-NP2, we tuned the algorithm hyperparameters of \Cref{tab:params} on a per-experiment basis.

\begin{table}
\centering
\begin{minipage}{1\linewidth}
        \resizebox{\linewidth}{!}{

\begin{tabular}{c l cccc | cccc | cccc | cccc}
\hline
\multirow{2}{*}{}&\multirow{2}{*}{Environment}& \multicolumn{4}{c}{Time [s]} & \multicolumn{4}{c}{Num Hyperplanes} & \multicolumn{4}{c}{Rel Volume} & \multicolumn{4}{c}{Frac in collision} \\
 &&NP& ZO & Greedy& Ray & NP & ZO& Greedy & Ray & NP& ZO & Greedy & Ray& NP& ZO & Greedy & Ray \\
\hline
\multirow{8}{*}{\rotatebox[]{90}{\textbf{Fast}}\;} 
&\texttt{Flipper} &0.168 & \textbf{0.025} & 0.045 & 0.847 & 22.7 & 15.8 & 13.6 & \textbf{13.4} & 1.0 & 0.627 & \textbf{2.167} & 2.017 & 0.040 & 0.013 & 0.022 & 0.016\\
&\texttt{UR3} &0.994 & \textbf{0.041} & 0.113 & 0.221 & 41.6 & 27.8 & 25.4 & \textbf{22.7} & 1.0 & 0.971 & \textbf{3.598} & 3.296 & 0.040 & 0.029 & 0.020 & 0.028\\
&\texttt{UR3Wrist} &1.227 & \textbf{0.050} & 0.168 & 0.948 & 51.4 & 34.0 & 30.8 & \textbf{26.5} & 1.0 & 3.824 & \textbf{5.376} & 1.663 & 0.029 & 0.031 & 0.023 & 0.023\\
&\texttt{IIWAShelf} &1.345 & \textbf{0.091} & 0.442 & 0.633 & 77.3 & 47.2 & 32.7 & \textbf{31.0} & 1.0 & 6.196 & \textbf{8.73e3} & 16.249 & 0.068 & 0.034 & 0.027 & 0.052\\
&\texttt{4Shelves} &1.317 & \textbf{0.092} & 0.313 & 0.845 & 62.9 & 46.9 & 35.1 & \textbf{30.3} & 1.0 & 0.139 & 1.388 & \textbf{1.724} & 0.102 & 0.038 & 0.025 & 0.033\\
&\texttt{IIWABins} &1.536 & \textbf{0.074} & 0.627 & 0.861 & 84.0 & 44.3 & 33.4 & \textbf{31.1} & 1.0 & 542.228 & \textbf{4.99e4} & 604.161 & 0.092 & 0.035 & 0.027 & 0.021\\
&\texttt{2IIWAs} &20.482 & \textbf{2.024} & 62.060 & 46.527 & 242.1 & 246.2 & 77.7 & \textbf{65.7} & 1.0 & 0.0008 & \textbf{316.777} & 95.052 & 0.061 & 0.057 & 0.037 & 0.055\\
&\texttt{Allegro} &2.277 & 0.264 & \textbf{0.101} & 0.106 & 75.3 & 83.7 & 41.9 & \textbf{39.6} & 1.0 & 0.036 & \textbf{152.040} & 60.554 & 0.0007 & 0.052 & 0.015 & 0.033\\
\hline
\multirow{8}{*}{\rotatebox[]{90}{\textbf{Precise}}\;} 
&\texttt{Flipper} &1.002 & 0.172 & \textbf{0.152} & 0.269 & 26.1 & 17.8 & \textbf{14.3} & 14.4 & 1.0 & 0.621 & 1.169 & \textbf{1.236} & 0.003 & 0.001 & 0.001 & 0.0004\\
&\texttt{UR3} &5.222 & \textbf{0.304} & 0.325 & 0.642 & 52.9 & 44.5 & 28.1 & \textbf{27.0} & 1.0 & 3.698 & \textbf{13.656} & 2.301 & 0.003 & 0.002 & 0.002 & 0.003\\
&\texttt{UR3Wrist} &4.664 & \textbf{0.359} & \textbf{0.359} & 0.410 & 61.8 & 56.1 & 31.5 & \textbf{28.6} & 1.0 & 3.811 & \textbf{183.345} & 4.959 & 0.002 & 0.003 & 0.001 & 0.002\\
&\texttt{IIWAShelf} &11.261 & \textbf{0.787} & 1.087 & 3.008 & 126.3 & 92.0 & 41.0 & \textbf{37.8} & 1.0 & \textbf{763.826} & 7.817 & 9.260 & 0.004 & 0.003 & 0.002 & 0.004\\
&\texttt{4Shelves} &25.315 & \textbf{0.699} & 0.838 & 1.002 & 85.2 & 81.3 & 40.5 & \textbf{36.1} & 1.0 & 0.124 & 1.592 & \textbf{1.672} & 0.012 & 0.003 & 0.002 & 0.003\\
&\texttt{IIWABins} &15.487 & \textbf{0.763} & 0.997 & 2.188 & 128.6 & 98.6 & 44.1 & \textbf{39.6} & 1.0 & 504.123 & \textbf{603.431} & 92.268 & 0.003 & 0.003 & 0.002 & 0.003\\
&\texttt{2IIWAs} &202.913 & \textbf{43.194} & 84.973 & 77.181 & 395.1 & 738.7 & 90.8 & \textbf{82.7} & 1.0 & 0.0003 & 37.507 & \textbf{48.984} & 0.006 & 0.005 & 0.004 & 0.004\\
&\texttt{Allegro} &2.164 & 7.455 & \textbf{0.499} & 1.271 & 75.3 & 290.6 & 45.5 & \textbf{42.2} & 1.0 & 0.005 & 8.237 & \textbf{20.434} & 0.0007 & 0.005 & 0.001 & 0.003\\
\hline
\end{tabular}
}
\end{minipage}
\vspace{0.2cm}
\caption{Statistics averaged over all 10 seed points.  ``Rel Volume'' corresponds to the volumes of the MVIE, normalized by that of IRIS-NP. For the ``Fast'' settings, on average, IRIS-ZO was 15.5 times faster than IRIS-NP with 1.4 times fewer hyperplanes. IRIS-NP2 with the greedy strategy was 6.55 times faster with 2.1 times fewer hyperplanes, and with the ray strategy was 4.2 times faster with 2.3 times fewer hyperplanes. For the ``Precise'' settings, IRIS-ZO was 14 times faster than IRIS-NP with around the same number of hyperplanes. IRIS-NP2 with the greedy strategy was 12.3 times faster with 2.4 times fewer hyperplanes, and with the ray strategy was 8 times faster with 2.7 times fewer hyperplanes.}
\label{tab:results}
\end{table}

\begin{figure}
\vspace{-0.5cm}
    \centering
    \includegraphics[width = \textwidth, trim = {0cm, 0cm, 0cm, 0cm}, clip]{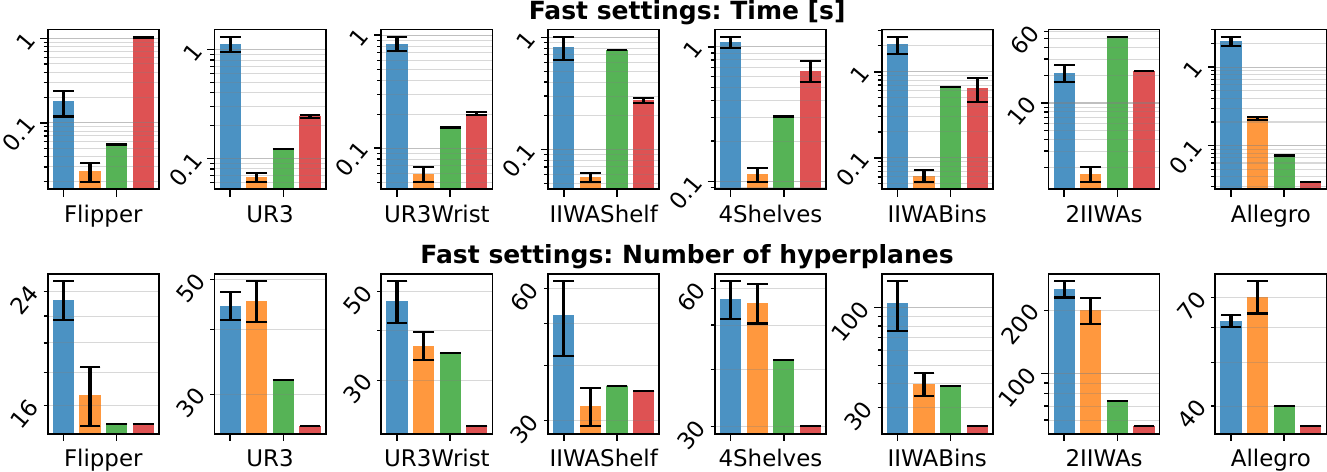}\\
    \includegraphics[width = \textwidth, trim = {0cm, 0cm, 0cm, 0cm}, clip]{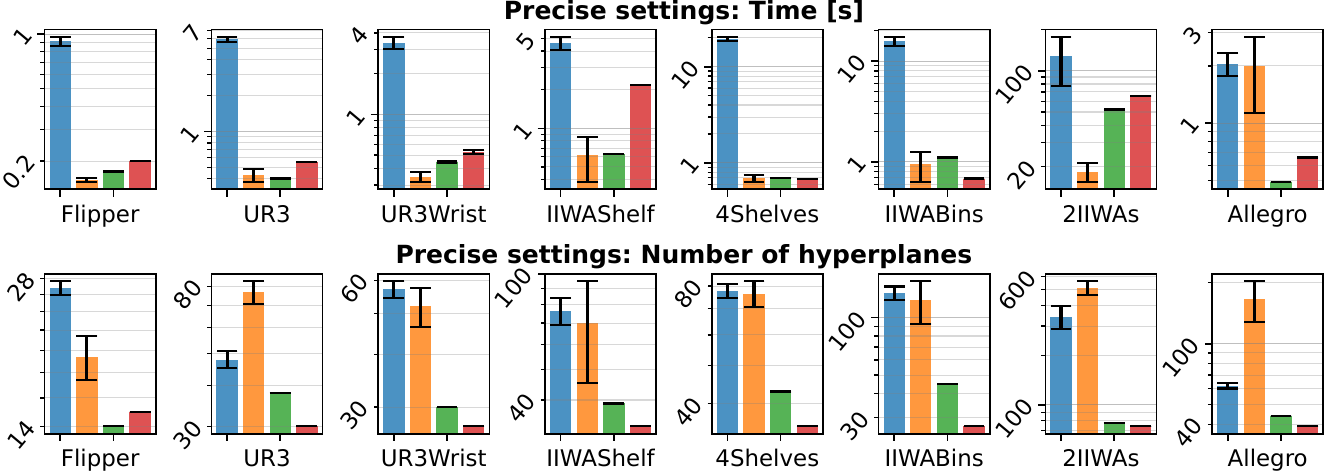}\\
    \includegraphics[width = \textwidth, trim = {0cm, .5cm, 0.5cm, 13cm}, clip]{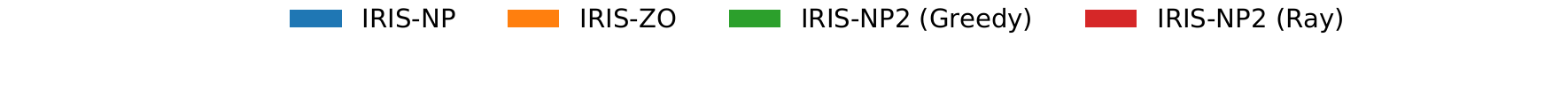}
    \vspace{-1cm}
    \caption{The mean and standard deviation of the runtime and hyperplane counts across ten trials per environment for the 8 seed configurations shown in \Cref{fig:environments}. }
    \label{fig:eval_fast}
\end{figure}

To mitigate the effects of randomness of the algorithms in our comparisons, we constructed regions around each seed point 10 times. This yields 100 trials per robot environment, or 800 total, for each algorithm. For the parallelized components of the algorithms, an Intel Core i9-10850K (10 cores, 20 threads) is used. All convex programs are solved using MOSEK \cite{mosek} and the nonlinear programs are solved using SNOPT \cite{gill2005snopt}.
The volume of the maximum-volume inscribed ellipsoid is used as a proxy for polytope volume. We report these volumes normalized by the averaged IRIS-NP results.

For each algorithm and robotic system, we include the runtime, number of hyperplanes, volume, and proportion of the region in collision, averaged over all trials on all seed points in \Cref{tab:results}.
Region volume and number of hyperplanes are both dependent on the seed configuration -- if the robot is in a narrow passageway, the region may be small, and if it is near obstacles, many hyperplanes may be required.
In \Cref{fig:eval_fast}, we visualize the runtime and hyperplane data across the 10 trials for the 8 corresponding seed configurations shown in \Cref{fig:environments}.  Our algorithms show significant improvements relative to IRIS-NP in runtime and number of hyperplanes.  In particular, IRIS-ZO gains approximately an order-of-magnitude time advantage with a comparable number of hyperplanes.  While IRIS-NP2 is slower than IRIS-ZO, it generally remains faster than IRIS-NP and uses significantly fewer hyperplanes.

\section{Discussion}\label{sec:discussion}
\looseness=-1
In recent years, region generation has posed a major roadblock for the adoption of new motion planning approaches such as \gls{gcs}. Toward alleviating this roadblock, we have presented significant improvements to the IRIS-NP algorithm, reducing its runtime, making it more user-friendly, and increasing the quality of the resulting regions. We derive a probabilistic test for the proportion of a given region that is collision-free, allowing the user to more directly trade off algorithmic precision and speed. We have also presented new approaches to the \textsc{SeparatingPlanes} subroutine, which show significantly better performance than IRIS-NP, while consistently achieving the user-specified collision-free threshold. IRIS-NP2 is generally faster and requires fewer hyperplanes to define the polytopes. IRIS-ZO is easy to implement is around 15 times faster and often uses fewer hyperplanes than IRIS-NP, although the regions are smaller. Furthermore, these new approaches scale more favorably with the complexity of the environment. As they are deployed in environments with increasingly many collision geometries, we anticipate the performance gap will grow even further.
Besides the quantitative improvements, the hyperparameters are straightforward to tune and have clear effects on the behavior of the algorithm.

\looseness=-1
There are myriad directions for future research. Better parallelization is a promising direction to speed up both new approaches -- beyond CPU-level multithreading, SIMD instructions~\cite{tan2019simdop} and GPU programming~\cite{lauterbach2009fast} have shown great promise for collision-checking. In particular, we regard better hardware usage for IRIS-ZO particularly promising, as both the sampling and the particle updates are trivially parallelizeable and currently bottle-necked by the number of threads in the CPU. Furthermore, using samples that are almost in-collision to initialize IRIS-NP2 might aid in finding small obstacles when requesting regions with a very small proportion in collision. Finally, we have focused on runtime and number of hyperplanes as our primary objectives, but more closely examining the relationship between the generated regions and the downstream motion planning algorithms will be essential for further improving the results. In particular, we aim to investigate how to improve efficient cover generation as in \cite{werner2024approximating}.

\vspace{-0.5cm}

\subsubsection{Acknowledgements} We thank Ravi Gondhalekar and Alexandre Amice for their feedback. This work was supported by Amazon.com, PO No. 2D-06310236, the MIT Quest for Intelligence, the Toyota Research Institute, and The Charles Stark Draper Laboratory, Inc., where Rebecca Jiang is a Draper Scholar and Ravi Gondhalekar is an employee.
\vspace{-0.5cm}

\bibliographystyle{IEEEtran}
\bibliography{biblio}
\end{document}